\documentclass{article}

\usepackage{microtype}
\usepackage{graphicx}
\usepackage{subfigure}
\usepackage{booktabs} %

\usepackage{hyperref}

\usepackage[accepted]{sysml2019}

\usepackage{enumitem}
\usepackage{mathptmx}
\usepackage{amsfonts}
\usepackage{listings}
\usepackage{xcolor}
\usepackage{multirow}
\usepackage{comment}
\usepackage{amsmath} 
\usepackage[capitalise]{cleveref} 

\definecolor{light-gray}{gray}{0.95}
\definecolor{mygreen}{rgb}{0,0.6,0}
\definecolor{unpred-color}{rgb}{0, 0.45, 0.69}
\definecolor{pred-color}{rgb}{0.87, 0.56, 0.19}
\definecolor{incorrect-color}{rgb}{0, 0.61, 0.45}

\begin{document}

\twocolumn[
\sysmltitle{Optimizing JPEG Quantization for Classification Networks}

\sysmlsetsymbol{equal}{*}

\begin{sysmlauthorlist}
\sysmlauthor{Zhijing Li}{cu}
\sysmlauthor{Christopher De Sa}{cu}
\sysmlauthor{Adrian Sampson}{cu}
\end{sysmlauthorlist}

\sysmlaffiliation{cu}{Cornell University}

\sysmlcorrespondingauthor{Zhijing Li}{zl679@cornell.edu}

\sysmlkeywords{Deep Learning, Deep Neural Network, Machine Learning, JPEG, Quantization table, Quantization Image Pipelining, Hyper-parameter Tuning}

\vskip 0.3in

\begin{abstract}
Deep learning for computer vision depends on lossy image compression:
it reduces the storage required for training and test data and
lowers transfer costs in deployment.
Mainstream datasets and imaging pipelines all rely on standard JPEG compression.
In JPEG, the degree of quantization of frequency coefficients controls the lossiness:
an 8$\times$8 \emph{quantization table (Q-table)} decides both the quality of the encoded image and the compression ratio.
While a long history of work has sought better Q-tables,
existing work either seeks to minimize image distortion or to optimize for models of the human visual system.
This work asks whether JPEG Q-tables exist that are ``better'' for specific vision networks and can offer better quality--size trade-offs than ones designed for human perception or minimal distortion.

We reconstruct an ImageNet test set with higher resolution to explore the effect of JPEG compression under novel Q-tables.
We attempt several approaches to tune a Q-table for a vision task.
We find that a simple \emph{sorted random} sampling method can exceed the performance of the standard JPEG Q-table.
We also use hyper-parameter tuning techniques including bounded random search, Bayesian optimization, and composite heuristic optimization methods.
The new Q-tables we obtained can improve the compression rate by $10\%$ to $200\%$ when the accuracy is fixed, or improve accuracy up to $2\%$ at the same compression rate.
\end{abstract}
]

\printAffiliationsAndNotice{}  %
\section{Introduction}
\label{sec:intro}
\begin{figure*}[t]
	\centering
	\includegraphics[width=0.8\linewidth]{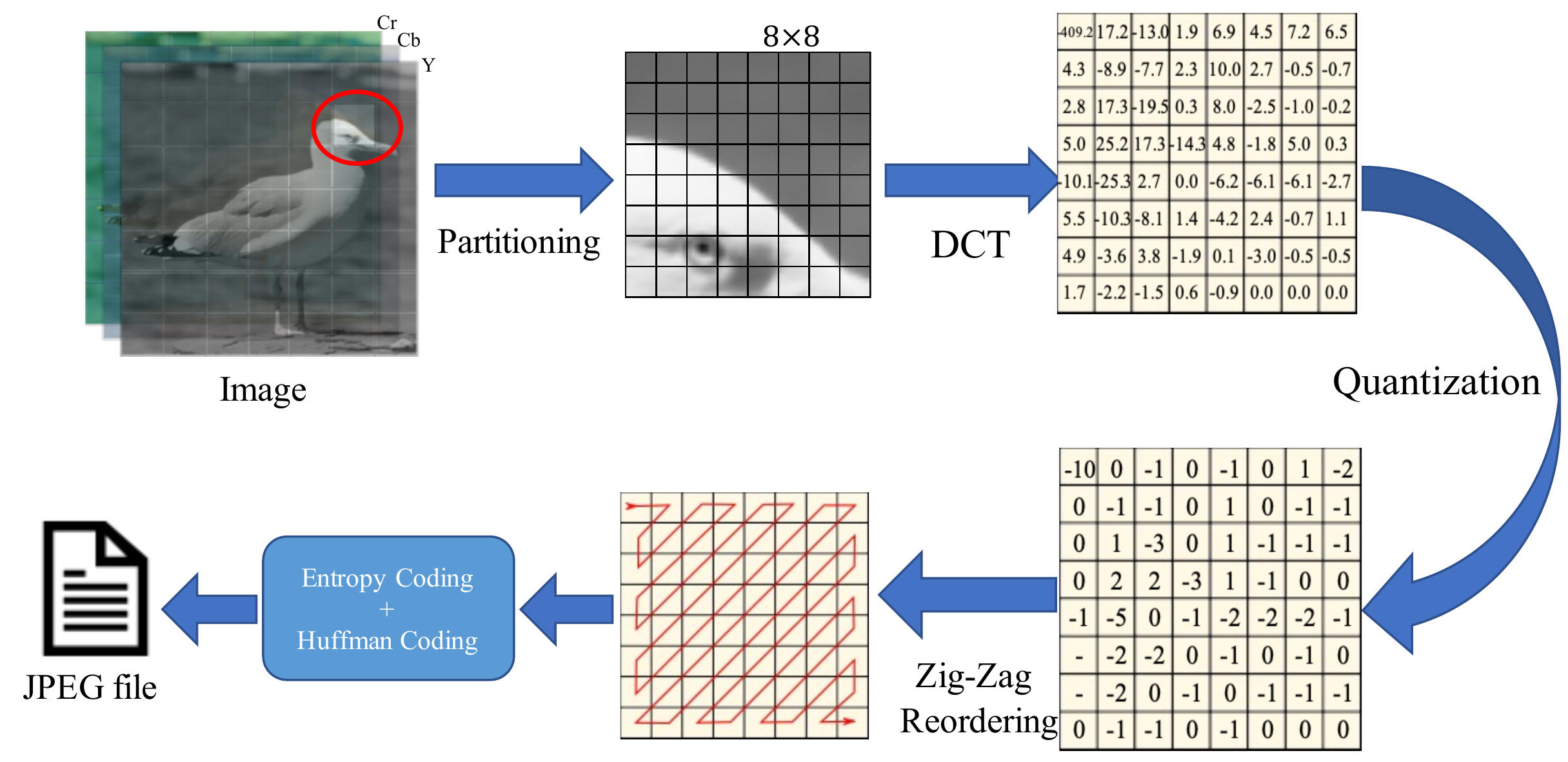}
	\caption{Overview of the JPEG compression algorithm.}
	\label{fig:jpeg_flow}
    \vspace{-2ex}
\end{figure*}
Deep neural networks for vision have extreme storage requirements.
For example, ImageNet is 150~GB~\cite{Deng2009ImageNetAL} and the Open Image Dataset requires 500~GB~\cite{kuznetsova2018open}.
If a model can achieve equivalent performance with images at a higher compression ratio, it can be cheaper to acquire, transmit, and store the necessary datasets.

Nearly all DNN datasets use the ubiquitous JPEG compression standard, including ImageNet~\cite{Deng2009ImageNetAL}, PASCAL VOC~\cite{pascal-voc-2012}, COCO~\cite{lin2014microsoft}, etc.
Because JPEG is a loss compression format, DNNs cope with some quality distortion in these images~\cite{dodge2016understanding}.
However, the JPEG standard is tuned for the human visual system (HVS) and preserves the components that are most relevant to human observers, rather than to computer vision algorithms.
This work asks whether machine learning pipelines would perform better if JPEG were instead tuned specifically for a given DNN.

Fig.~\ref{fig:jpeg_flow} shows an overview of the standard JPEG compression algorithm. Each spatial component (i.e., each YCbCr channel) is first partitioned into 8$\times$8 non-overlapping blocks. Then the spatial components are transformed to frequency components with the 2D discrete cosine transform (DCT).
Next, the algorithm quantizes the frequency components using a quantization table or \emph{Q-table}.
The Q-table's scale is determined by the \textit{quality factor} ranging from 1 to 100~\cite{LuaDist_2015}. For instance, a quality factor of 100 scales Q-table coefficients to 0 and the frequency components are rounded to integers. Then, the quantized coefficients are serialized using a \textit{zig-zag} ordering and losslessly compressed with entropy coding and Huffman coding.

Quantization is the lossy part of the algorithm. The Q-table decides how many bits to use for each frequency bucket, potentially playing a critical role for DNNs that use JPEG-compressed images.
However, existing research on designing Q-tables mostly prioritizes human-precepible distortion as the compression target~\cite{liu2018deepn}, which may compress and quantize features important to DNNs~\cite{4483511}.
Prior work instead targets DNN accuracy~\cite{liu2018deepn}, but it only considers \emph{recompressing} existing JPEG-compressed datasets rather than changing the way data is compressed in the first place.

In this work, use a high-resolution dataset to simulate the effect of compressing raw pixels with novel, task-specific quantization.
We use a range of optimization methods to tune a customized Q-table for a specific DNN that maximizes both vision task accuracy and compression ratio.
Starting with a simple baseline method, \emph{sorted random search}, we find Q-tables that improve ResNet50 top-1 accuracy by 1--2\% at the same compression rate as standard JPEG or improve compression rate by 20--200\% at the same accuracy.
We then use a range of hyper-parameter tuning methods, including Bayesian optimization and a composite heuristic autotuner, to improve on these Q-tables within bounds derived from the best results from the baseline method.
We find that these autotuning approaches yield small improvements over the simple approach, but after cross validation, the methods perform similarly.
Surprisingly, our simple sorted random search approach performs as well as the more advanced tuning approaches, and the advantages over standard JPEG are statistically significiant.

The contributions of this paper are:
\begin{itemize}[noitemsep]
\item We construct a high-resolution dataset for experimenting with novel compression techniques based on ImageNetV2~\cite{recht2019imagenetv2}.
\item We find that there is an opportunity to save space for the same accuracy by customizing the JPEG quantization table.
\item In particular, we recommend using a sorted random search procedure, which quickly produces tables that meet and exceed the standard JPEG quantization tables.
\item Despite extensive experimentation with other optimization procedures, we did not find a method that obviously outperforms sorted random search.
\end{itemize}
\vspace*{-\baselineskip}

\section{Related Work and Motivation}
\label{sec:relatedwork}
\subsection{Q-Table Optimization}
Because JPEG is one of the most widely used image compression techniques, JPEG Q-table optimization has been an enduring issue. Existing approaches to design JPEG Q-tables aim to maximize image quality in one of two ways: by minimizing a simple distortion metric such as PSNR~\cite{wu1993rate,fung1995design,ramchandran1994rate}
or by using a model of the human visual system (HVS)~\cite{watson1993dctune, wang2001designing, westen1996optimization, jiang2011jpeg}.
The HVS approach aims at optimizing visual image quality, where high-frequency components are viewed as less important than low-frequency components. Standard JPEG is designed with the same goal~\cite{wallace1992jpeg}.
This work's hypothesis is that the best Q-tables for DNN vision may be different than ones that emphasize visual quality because of different sensitivity to frequency components~\cite{liu2018deepn}.

\subsection{DeepN-JPEG}
There is some work that focuses on the optimization of Q-tables for DNNs.
DeepN-JPEG~\cite{liu2018deepn} exploits the difference between the HVS and DNNs and
designs a Q-table according to the frequency bands.
The Q-table search space is limited by design: the frequency bands share a small number of parameters.
We explore a larger space using hyper-parameter tuning.
Also, DeepN-JPEG both tunes and tests on ImageNet, without cross validation on another dataset.
ImageNet is a low-resolution, compressed dataset, and that work \emph{recompresses} it.
We instead focus on simulating compression of raw pixel data using a high-resolution rebuild from
ImageNetV2~\cite{recht2019imagenetv2} and validate our result on different datasets to check robustness.

\section{Vision Task}
\label{sec:setting}
For the purpose of tuning Q-tables for DNNs, we need to choose a particular vision task. We use a classification network, ResNet50~\cite{he2016deep}, and reconstruct a dataset to simulate raw pixel data.

\subsection{Constructing a Dataset}
\label{sec:dataset}
ImageNet is already downsized and lossily compressed. For instance, the ImageNet 2013 classification dataset has an average resolution of 482$\times$415 pixels~\cite{russakovsky2015imagenet}.
Instead, we turn to ImageNetV2, a new test dataset for ImageNet~\cite{recht2019imagenetv2}. In ImageNetV2, the images are also downsized and compressed, but the authors have open-sourced the code to generate the dataset as well as the IDs that let us re-download the original, large images from Flickr~\cite{Flickr}.
The large images are also JPEG-compressed, but they have an average resolution of 1933$\times$1592 pixels. We then downsize them to a smaller size compatible with ImageNet. The resizing removes JPEG compression artifacts and simulates uncompressed images at those sizes.

ImageNetV2 contains three test sets representing different sampling strategies: MatchedFrequency, TopImages, and Threshold0.7.
Each test set has 1000 classes with 10 images per class. Specifically, we use the MatchedFrequency as our training dataset, which is sampled to match the MTurk selection frequency distribution of the original ImageNet validation set for each class.

\subsection{Tuning Methodology}
\label{sec:train-setup}
Hyper-parameter tuning for all kinds of DNNs is unrealistic.
We focus on one particular DNN:
ResNet50~\cite{he2016deep} as implemented in PyTorch~\cite{paszke2017automatic}.
We use 500 classes with 5 images each from ImageNetV2's MatchedFrequency dataset to speed up compression. The optimization targets are top-1 accuracy and the compression rate, calculated as the ratio of the size of raw bitmap data to compressed JPEG images.

\section{Sorted Random Search}
\label{sec:sorted}

We start with a simple method that randomly samples Q-tables.
The idea is to get a baseline to see how hard it is to find, among the space of all Q-tables, new ones that mimic the performance of the standard JPEG tables---or that outperform standard JPEG \emph{for a specific DNN for a specific vision task}.

\subsection{Method}
The search space for a uniform random search is $256^{64} \approx 1.34\cdot 10^{154}$. We need to decrease our search space.
Lower frequencies probably matter more than higher frequencies. That is why the upper left of the standard table are large and the lower right are small, as shown in Fig.~\ref{fig:jpeg_flow}. A simple sorting strategy can restrict sampling to tables that preserve this ordering. We generate 64 random numbers in the range $[s, e]$, where $ s, e \in \mathbb{Z} $ and $1 \le s<e \le 255 $, sort them, and build an 8$\times$8 table using the zig-zag ordering.
We refer to this Q-table sampling strategy as \textit{sorted random search}. 
\subsection{Performance}
\label{sec:sorted-performance}
\begin{figure*}[t]
\centering
\begin{minipage}[]{.31\textwidth}
    \includegraphics[width=\linewidth]{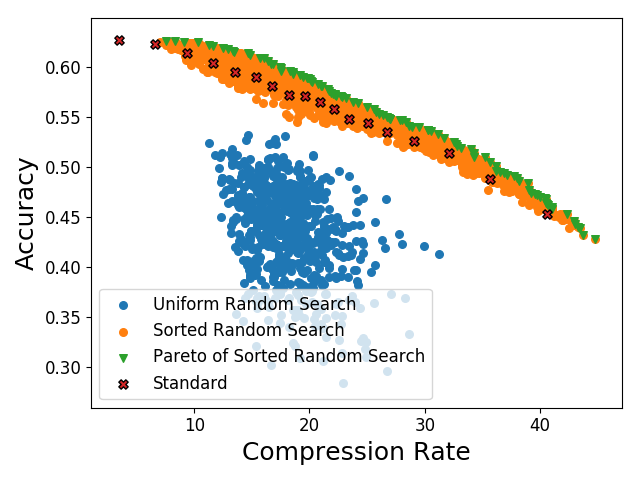}
    \vspace*{-8mm}
\caption{Sorted random search and random search performances compared to standard JPEG.}\label{fig:st_vs_sorted}
\end{minipage}\hfill
\begin{minipage}[]{.31\textwidth}
    \includegraphics[width=\linewidth]{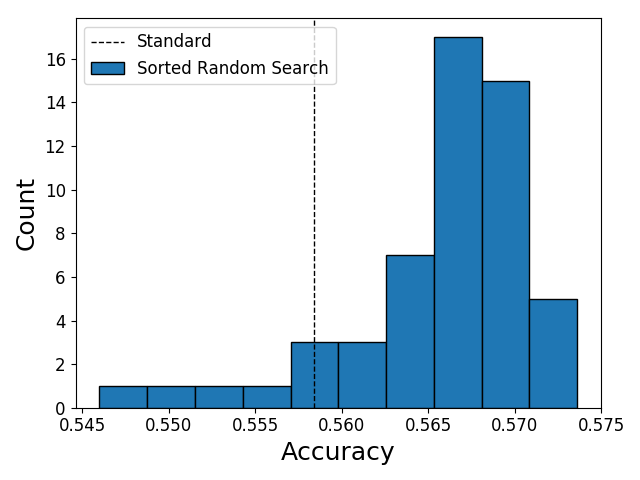}
    \vspace*{-8mm}
\caption{Sorted random search distribution with compression rate range $(22,22.2)$.}\label{fig:sorted_distribution}
\end{minipage}\hfill
\begin{minipage}[]{.31\textwidth}
    \includegraphics[width=\linewidth]{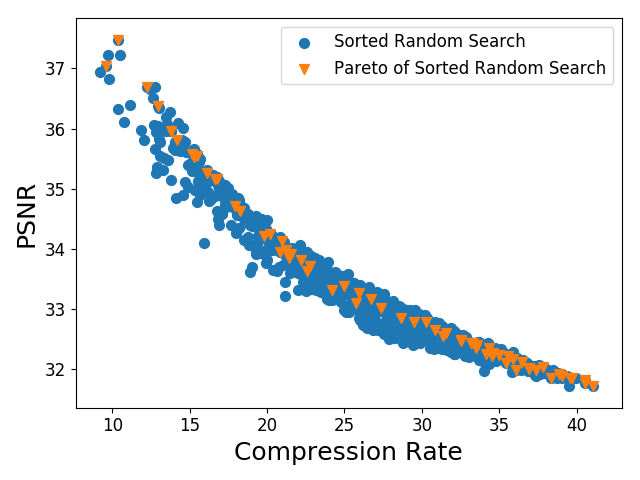}
    \vspace*{-8mm}
\caption{PSNR of sorted random search with Pareto-optimal points according to DNN accuracy.}\label{fig:psnr}
\end{minipage}
\begin{minipage}[]{.5\textwidth}
\centering
    \includegraphics[width=0.62\linewidth]{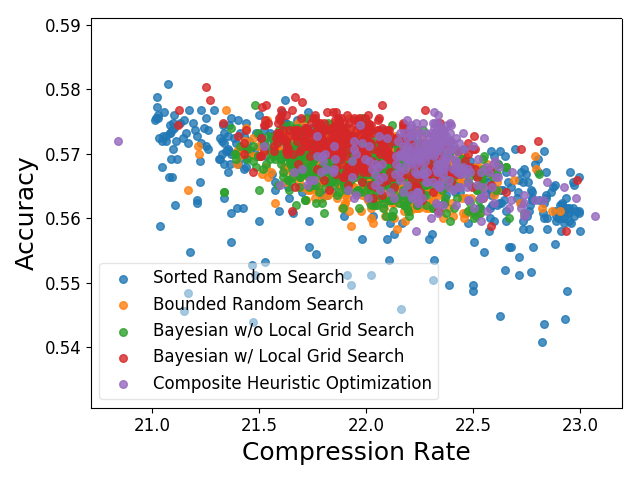}
    \vspace*{-6mm}
\caption{Bounded Searching Methods}\label{fig:bound_algs}
\end{minipage}\hfill
\begin{minipage}[]{.5\textwidth}
\centering
    \includegraphics[width=0.62\linewidth]{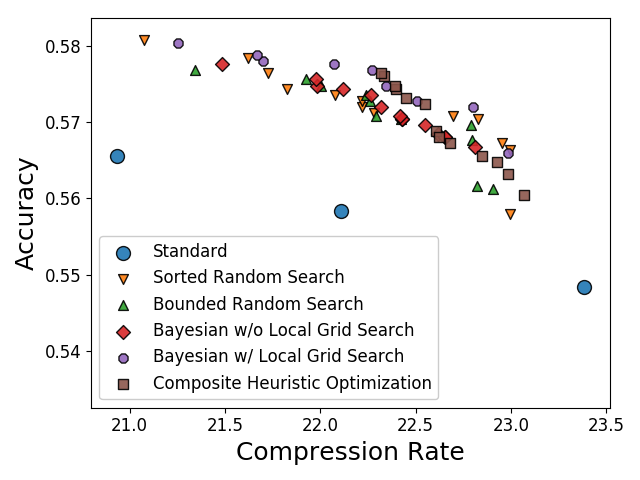}
    \vspace*{-6mm}
\caption{Pareto of bounded searching methods.}\label{fig:bound_pareto}
\end{minipage}
\end{figure*}
Fig.~\ref{fig:st_vs_sorted} compares 4000 Q-tables sampled using sorted random search, 1000 from uniform random search, and the standard Q-tables with quality factor from 10 to 100 at an interval of 5 in terms of compression rate and accuracy. We also plot the Pareto frontier for sorted random search. With uniform random search, we cannot find any Q-table that outperforms the standard JPEG Q-table. Sorted random search finds Pareto-optimal points that outperform standard JPEG by 1\% to 2\% in top-1 accuracy given the same compression rate when the quality factor of standard JPEG is in the range $[15,90]$. 
Given the same accuracy, sorted random search can find Q-tables with compression rates 10\% to 200\% higher than standard JPEG depending on the baseline compression rate. 
We further take a close look at points with closest compression rate and accuracy to the quality-50 Q-table of standard JPEG, where the standard table is scaled at 100\%. The improvements of accuracy and compression rate are 1.5\% and 12.9\%.

Fig.~\ref{fig:sorted_distribution} shows the distribution of sorted random points with compression rate between 22 to 22.2. Among 52 points, there are 48 points that perform at least as well as the standard JPEG Q-table, of which 93.75\% perform better. %
Taking into account the strong randomness of sorted random search, the result indicates sorted random search can easily find many Q-tables that outperform standard JPEG.

To demonstrate the difference between the PSNR and DNN optimization goals, Fig.~\ref{fig:psnr} plots the PSNR of sorted random search points and we highlight the Pareto frontier in terms of compression rate and accuracy. Although PSNR is closely related to DNN accuracy, it cannot be taken as the sole indicator of high accuracy.

\begin{figure*}
\centering
\begin{minipage}[]{.32\textwidth}
 \includegraphics[width=\linewidth]{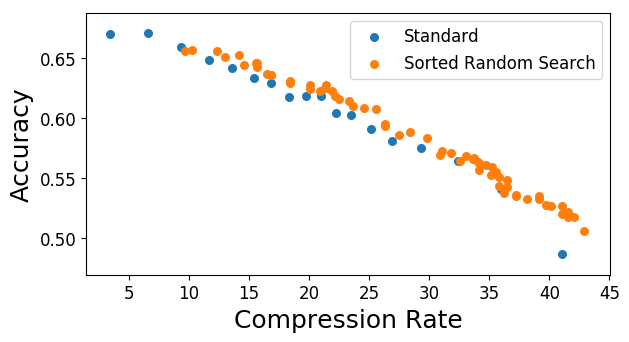}\vspace*{-1mm}
    \includegraphics[width=\linewidth]{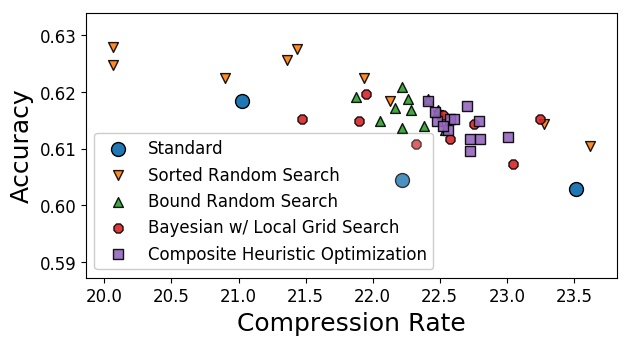}
    \vspace*{-8mm}
\caption{Pareto performance validated on MatchedFrequency (part) in ImageNetV2.}\label{fig:cv_mf}
\end{minipage}\hfill
\begin{minipage}[]{.31\textwidth}
	\includegraphics[width=\linewidth]{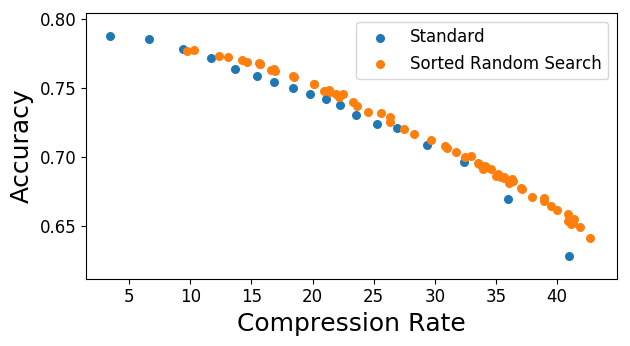}\vspace*{-1mm}
    \includegraphics[width=\linewidth]{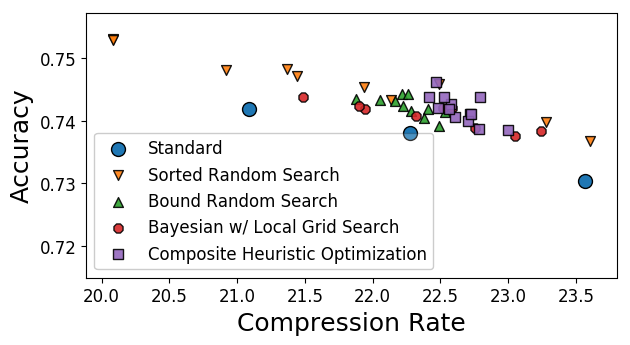}
    \vspace*{-8mm}
\caption{Pareto performance validated on TopImages in ImageNetV2.}\label{fig:cv_ti}
\end{minipage}
\hfill
\begin{minipage}[]{.31\textwidth}
	\includegraphics[width=\linewidth]{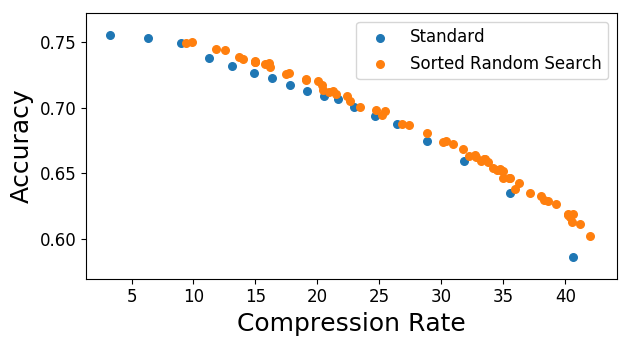}\vspace*{-1mm}
    \includegraphics[width=\linewidth]{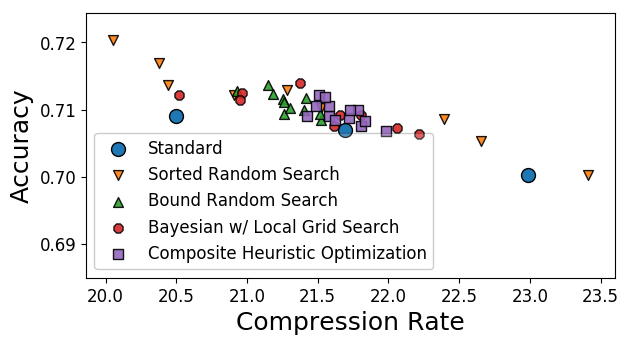}
    \vspace*{-8mm}
\caption{Pareto performance validated on ImageNet validation set.}\label{fig:cv_imgnt}
\end{minipage}
\end{figure*}

\section{Bounded Search}
\label{sec:bound}

The random search method from the previous section
samples from a tremendously large search space but uses an ordering constraint to make the search tractable.
In this section, we hypothesize that, if we limit our search space of Q-table components to a smaller range, we can eliminate ordering constraint and find better tables using more sophisticated optimization methods.

To limit the search space, we need to focus on a particular compression rate range.
We focus on a regime close to standard JPEG at quality 50, where the compression rate is approximately 22.
Let $P$ be the set of Q-tables on the Pareto frontier and $P' \subseteq P$ be the set of Q-tables with compression rate ranges in $[21,23]$.
Let $P''$ be the set of Q-tables in $P'$ and their transpose. Let $M_{P''}$ be the $n \times 8 \times 8$ array that results from concatenating the Q-tables in $P''$. We can get a symmetric upper bound matrix by taking maximum along first dimension of $M_{P''}$. Therefore, the $8\times 8$ boundary matrices are set to:
\begin{align*}
M_{\text{Lower Bound}}[j,k] =&  \min_i(M_{P''}[i,j,k])
- 0.5 \sigma_i (M_{P''}[i,j,k])\\
M_{\text{Upper Bound}}[j,k] =&  \max_i(M_{P''}[i,j,k])
+ 0.5 \sigma_i (M_{P''}[i,j,k])
\end{align*}
where $i\in[0,n-1], j\in[0,7], k\in[0,7]$ and $\sigma_i$ takes the standard deviation along first dimension of a matrix.
We can sample Q-tables with the two boundary matrices to limit the exploration space.
We expect that Q-tables sampled within these boundaries will result in compression rates in the range $[21,23]$.

\subsection{Bounded Random Search}
The most intuitive way as a baseline for bounded search is uniformly sampling Q-tables within the boundaries, which we called \textit{bounded random search}. 
We plot Q-tables sampled using bounded random search in Fig.~\ref{fig:bound_algs} and corresponding Pareto frontiers in Fig.~\ref{fig:bound_pareto}.

\subsection{Bayesian Optimization}
Bayesian optimization is widely used when evaluation of the objective function is expensive. It approaches the objective function with a surrogate model, which is cheap to evaluate.
To take the advantage of Bayesian optimization, there are two problems we need to address.

First, Bayesian optimization optimizes a single target objective function and our task involves two: compression rate $CR$ and accuracy $ACC$. We therefore use a parabola $\text{fitness}(CR) = a CR^2 + b CR + c$ to fit the Pareto frontier formed by Q-tables obtained in sorted random search. Then we can set the target value $y = ACC-\text{fitness}(CR)$, standing for the difference between the accuracy gap between actual accuracy and best accuracy found by sorted random search.
\begin{figure}[t]
	\centering
	\includegraphics[width=0.55\columnwidth]{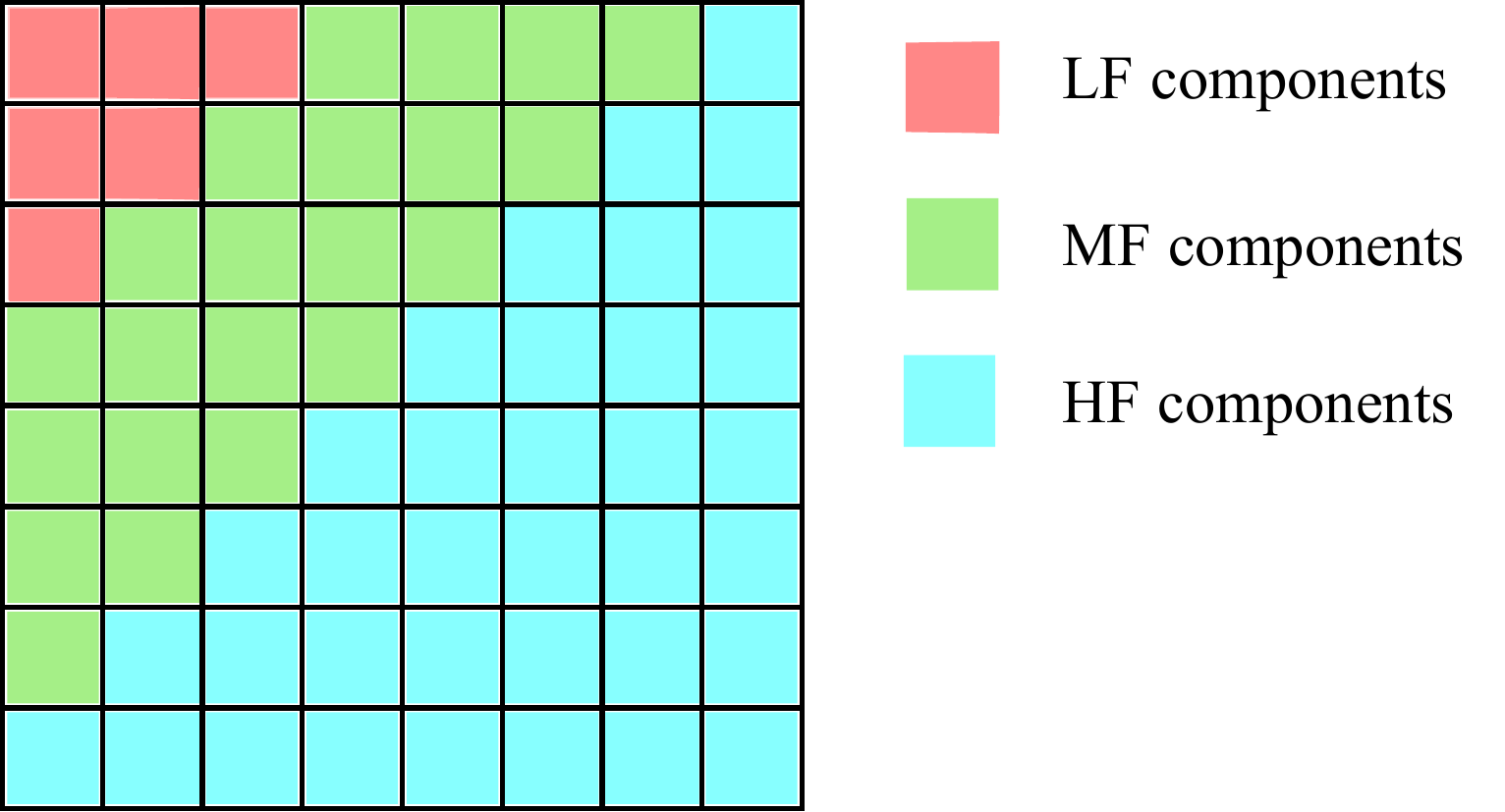}
	\caption{Frequency band partitions.}
	\label{fig:freq_band}
\end{figure}

Second, the search space is huge---as large as $1.34\cdot 10^{154}$.
We therefore set the sampling bound the same as the one in the previous section. Also, the low frequency (LF) and middle frequency (MF) components of DCT 8$\times$8 arrays have larger absolute values~\cite{kaur2011steganographic}, and therefore are more susceptible to changes in quantization values. We define the \textit{area of interest} as the low frequency and middle frequency bands as shown in Fig~\ref{fig:freq_band}. We first generate $10^5$ random Q-tables and keep the one with highest acquisition function value. Then we randomly choose 5 indices of that Q-table in the area of interest and perform a local grid search that exhaustively list possible Q-tables within the bound. The Q-table with highest acquisition function value is kept. The process is repeated 20 times.
In our experiments, we try Bayesian optimization both with and without local random search for comparison and plot them in \cref{fig:bound_algs,fig:bound_pareto}.
\subsection{Composite Heuristic Optimization}

Different search methods have different strengths.
We next apply
OpenTuner~\cite{ansel:pact:2014},
a generic auto-tuning framework that uses
a \emph{multi-armed bandit} (MAB) approach to select between several methods.
OpenTuner maximizes an objective by picking a strategy among different optimization approaches (the bandit \emph{arms}).
We select the
particle swarm optimization (PSO), simulated annealing (SA), differential evolution (DE), greedy mutation, and random Nelder--Mead heuristics as the bandit arms.
All these heuristics are generic optimization algorithms and can be performed without much computation overhead.

\cref{fig:bound_algs,fig:bound_pareto} show that this method produces results that perform as successfully as Bayesian optimization.

\subsection{Performance}
Fig.~\ref{fig:bound_algs} shows
618 points from bounded random search, 610 points from Bayesian optimization without local grid search, 416 points from Bayesian optimization with local grid search, and 262 points from composite heuristic optimization.
We also plot the points from sorted random search in the range $[21,23]$ as a baseline. Fig.~\ref{fig:bound_pareto} shows that the %
Pareto frontiers of all methods and standard JPEG. All proposed methods can find Q-tables that outperform standard JPEG. Simple techniques like bounded random search can give a better distribution than sorted random search, but the Pareto frontier does not improve. Bayesian optimization without local grid search does not show a strong advantage over sorted and bounded random search mainly because the sampling space is too large to find a promising point to probe. Bayesian optimization with local grid search and composite heuristic optimization are the two methods with the highest means, smallest standard deviations, and the best Pareto frontiers. Given the same compression rate, the accuracy improvement can be up to 2.5\% over standard JPEG. In particular, the point with the closest compression rate to quality-50 standard JPEG is found by Bayesian optimization, and the absolute accuracy improvement is 1.9\%.

\section{Evaluation}
\label{sec:evaluation}
This section examines accuracy under cross validation and the efficiency of each method.

\subsection{Cross Validation}
\label{sec:generalize}
The experiments above find good Q-tables using part of ImageNetV2's MatchedFrequency dataset.
It is important to check whether these Q-tables perform consistently on other datasets. We cross-validate our Q-tables on the other 500 classes of the reconstructed MatchedFrequency dataset with 5 images each in Fig.~\ref{fig:cv_mf}, the reconstructed TopImage dataset of ImageNetV2 in Fig.~\ref{fig:cv_ti}, and the original ImageNet dataset with 10 images per class in Fig.~\ref{fig:cv_mf}.

The Pareto frontier of sorted random search becomes less smooth. 
When the compression rate is the same, the gap of accuracy improvement between our methods and standard JPEG decreases to at most 1.5\% for the MatchedFrequency and TopImages datasets (\cref{fig:cv_mf,fig:cv_ti}) and 1\% for ImageNet (\cref{fig:cv_imgnt}) but never disappears. Bayesian optimization and MAB no longer significantly outperform sorted random and bounded random search. Overall, after cross validation, the complex algorithms do not show any advantage over our simple baseline algorithm, sorted random search.
\subsection{Significance of Improvement}
\label{sec:significance}
\begin{table}[t]
\caption{Significance of improvement over baseline, where * denotes mean differences that are statistically significant according to a Student $t$-test with $p < 10^{-5}$ and ** denotes $p < 10^{-11}$.}
\vspace{-2ex}
\label{tab:significance}
\begin{center}
\begin{small}
\begin{sc}
\begin{tabular}{lccr}
\toprule
  & \begin{tabular}[c]{@{}l@{}}Matched \\Frequency\end{tabular} & ImageNet\\
\midrule
Sorted Random Search
& 0.91\%** & 1.16\%**
\\
Bounded Random Search 
& 0.72\%*  & 0.66\%*
\\
\begin{tabular}[c]{@{}l@{}}Bayesian Optimization \\ w/ Local Grid Search\end{tabular} 
& 0.55\%*  & 0.77\%*
\\
\begin{tabular}[c]{@{}l@{}}Composite Heuristic \\Optimization\end{tabular}                                                                                          & 0.73\%*  & 1.17\%**\\
\bottomrule
\end{tabular}
\end{sc}
\end{small}
\end{center}
\end{table}
To check whether the improvement of our methods over standard JPEG is significant, we perform statistical analysis of accuracy improvement given the closest compression rate. We test one Q-table obtained by each method against standard JPEG with quality set to 50. To avoid unintentionally cheating, we choose the point with the closest compression rate that is \textit{larger} than that for the quality-50 standard Q-table, i.e., we allow accuracy to be a little worse. For each Q-table, we randomly sample 100 datasets from the original ones with 700 classes and 4 images per class. Then we compare the difference in mean accuracy and check significance using Student's $t$-test.

Table~\ref{tab:significance} shows the difference of mean top-1 accuracy between standard JPEG and our Q-table obtained by different methods, on the MatchedFrequency and ImageNet Validation dataset. All methods find Q-tables with statistically significant accuracy improvement.

\subsection{Efficiency}
\label{sec:efficiency}
Dataset compression is expensive.
For instance, the time to compress our training dataset and measure accuracy is approximately 3 minutes. A good method should find optimal Q-tables in as few trials as possible and the decision time for each trial should be short.

Table~\ref{tab:efficiency} shows the results of profiling each method on a server with two Intel Xeon(R) E5-2620 v4 CPUs with 8 cores per socket and 2 threads per core.
The table
lists two factors affecting efficiency: the decision time for an algorithm to decide the next point to search, and the number of trials required to generate the first 10 good points.
``Good'' points are defined as those with $\text{Acc} - \text{fitness}(\text{CR}) > -0.001$.
The decision time is given as the average over 100 trials, rounded to two significant digits.
Among all methods, composite heuristic optimization requires the fewest trials to generate good points, while bounded random search takes the most trials to find good points. The decision time for all except Bayesian optimization is negligible, as the latter requires an expensive local grid search. %
\begin{table}[t]
\caption{Efficiency of different methods.}
\vspace{-2ex}
\label{tab:efficiency}
\begin{center}
\begin{small}
\begin{sc}
\begin{tabular}{lccr}
\toprule
Method & 
\begin{tabular}[c]{@{}l@{}}Decision\\Time\end{tabular} & 
\begin{tabular}[c]{@{}l@{}}Trials for\\10 Good Points\end{tabular}
\\\midrule
Sorted random search & 1.2ms                                                                              & 412                                                                                                    \\
Bounded random search                      &  12.ms                                                                               & 617                                                                                                    \\
\begin{tabular}[c]{@{}l@{}}Bayesian optimization\\w/ local grid search\end{tabular} & 60.s                                             & 253                                                                                                    \\
\begin{tabular}[c]{@{}l@{}}Composite Heuristic \\Optimization\end{tabular} & 5.1ms                                                        & 150                                                                                                  \\
\bottomrule
\end{tabular}
\end{sc}
\end{small}
\end{center}
\vskip -0.2in
\end{table}

\section{Conclusion and Future Work}
Our research shows the potential for accuracy and compression rate improvement by redesigning the JPEG Q-table, a topic that lacks attention from researchers in both of the image compression and DNN communities. All our proposed methods obtain better Q-tables compared to the standard one with an accuracy gain of $1\%$ to $2\%$ when the compression rate is fixed and a compression rate increase of 10\% to $200\%$ given the same accuracy.
Although a composite heuristic optimizer stands out in efficiency and accuracy, we recommend a simple sampling technique, \emph{sorted random search}, which can find good quantization tables at a wide range of compression--accuracy trade-offs.
The resulting improved Q-tables outperform standard JPEG for our vision task in cross validation.

In this work, we consider the DNN fixed and only tune compression. But the network was trained on data using standard JPEG and therefore might perform worse on images using our new Q-tables. Fine-tuning retraining might help adjust the networks to match the new image characteristics, further improving the accuracy with nonstandard quantization.
We have done a preliminary retraining experiment, which shows a positive result.
Avenues for future work include further work on DNN retraining and applying our techniques to
more vision applications such as object detection, semantic segmentation, and other tasks.

\section*{Acknowledgments}
This research was supported by gifts from Google and Nvidia.
We thank our colleagues Mark Buckler, Philip Bedoukian, and Jenna Choi, who initiated this research direction.
We also thank Zhiyong Hao and Danna Ma for comments that improved the manuscript.

\bibliography{references}
\bibliographystyle{sysml2019}

\end{document}